\documentclass{article}

\usepackage{arxiv}

\usepackage[utf8]{inputenc} % allow utf-8 input
\usepackage[T1]{fontenc}    % use 8-bit T1 fonts
\usepackage{hyperref}       % hyperlinks
\usepackage{url}            % simple URL typesetting
\usepackage{booktabs}       % professional-quality tables
\usepackage{amsfonts}       % blackboard math symbols
\usepackage{nicefrac}       % compact symbols for 1/2, etc.
\usepackage{microtype}      % microtypography
\usepackage{lipsum}
\usepackage{graphicx}
\graphicspath{ {./images/} }

\usepackage{xcolor}
\usepackage{amssymb}
\usepackage{cite}
\usepackage{color, colortbl}
\usepackage{subfigure}

\definecolor{c_green}{RGB}{230,255,230}
\definecolor{c_red}{RGB}{255,230,230}

\title{Intersection over Union with smoothing \\for bounding box regression}

\author{
 Petra \v Stevuli\'akov\'a \\
  University of Ostrava, Centre of Excellence IT4Innovations\\
  Institute for Research and Applications of Fuzzy Modeling\\
  30. dubna 22, Ostrava,  Czech Republic\\
  \texttt{Petra.Stevuliakova@osu.cz} \\
   \And
 Petr Hurtik \\
  University of Ostrava, Centre of Excellence IT4Innovations\\
  Institute for Research and Applications of Fuzzy Modeling\\
  30. dubna 22, Ostrava,  Czech Republic\\
  \texttt{Petr.Hurtik@osu.cz} \\
}

\begin{document}
\maketitle
\begin{abstract}
We focus on the construction of a loss function for the boun\-ding box regression. The Intersection over Union (IoU) metric is improved to converge faster, to make the surface of the loss function smooth and continuous over the whole searched space, and to reach a more precise approximation of the labels. The main principle is adding a smoothing part to the original IoU, where the smoothing part is given by a linear space with values that increases from the ground truth bounding box to the border of the input image, and thus covers the whole spatial search space. We show the motivation and formalism behind this loss function and experimentally prove that it outperforms IoU, DIoU, CIoU, and SIoU by a large margin. We experimentally show that the proposed loss function is robust with respect to the noise in the dimension of ground truth bounding boxes. The reference implementation is available at gitlab.com/irafm-ai/smoothing-iou.
\end{abstract}

% keywords can be removed

\begin{figure}[!h]
    \centering
    ~~~~~~~~~IoU~~~~~~~~~~~~~~~~~~~~~~DIoU~~~~~~~~~~~~~~~~~IoU+smooth\\
    \includegraphics[width=81mm]{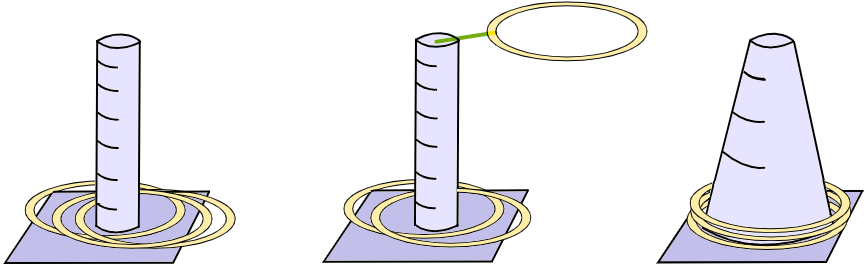}
    \caption{An illustration of the proposed loss given by a child game of throwing rings to a stick. The ring and the stick represent the predicted box and the loss function, respectively. The goal is to place the ring in order to centre it with respect to the stick. The standard IoU loss represents the simple stick, the DIoU loss uses a rubber between the ring and the stick to move (converge) it faster. The proposed IoU with smoothing is represented by a cone covering the loss space and continuously navigating the ring.} 
    \label{fig-intuition}
\end{figure}

\section{Problem formulation}
Object detection is an essential part of computer vision and is presented in areas of (identity) object tracking~\cite{pal2021deep}, optical quality evaluation~\cite{lei2022precise}, or autonomous driving~\cite{li2022cross} to name a few. The current state of the art is exclusively given by data-driven approaches, i.e., deep neural networks (convolutional or transformer-based types are used the most) that replaced older model-driven methods that suffered for precision and robustness.
The detection itself consists of three parts: confidence regression, object classification, and bounding box regression; regardless, we speak about one-step~\cite{redmon2018yolov3} or two-step~\cite{RenFasterRCNN} deep learning approaches. The three mentioned parts appear in a compound loss function and thus are critical for the training of a neural network.  

In this paper, we focus on the construction of a new loss function for the bounding box regression. The current way of research improves the well-known Intersection over Union (IoU) metric to converge faster, to make the surface of the loss function smooth and continuous, and to reach a more precise approximation of the labels. Similarly, our approach aims at these aspects but is motivated by the data-centric~\cite{jarrahi2022principles} approach: we assume that the training data set is small and that some labels are not perfect, so \emph{the loss must converge efficiently and be noise-robust}. Note that these are the requirements that naturally appear during cooperation with industry sector.

Here, we propose to enrich the standard IoU loss function with a smoothing part motivated by label smoothing~\cite{muller2019does}, whose purpose is twofold. Firstly, it guides the positioning through the whole domain (image) and, secondly, it weakens the effect of noisy labels. See Figure~\ref{fig-intuition} that shows the intuition behind the proposed approach. We show the motivation and formalism behind it in Section~\ref{sec-smooth} and demonstrate in Section~\ref{sec-benchmark} that it outperforms the other IoU variants by a large margin. Noise robustness is benchmarked up to noise of 60\% of the side size of the bounding box and shows that the decrease in test accuracy is minor. This is a valuable property because the integration of the loss is simple and does not require changes in the architecture compared to the teacher/student scheme that is commonly used when such robustness is required.

\section{Related work and preliminaries}
Currently, most of the loss functions for the bounding box regression fall into two categories: $\ell_n$-norm losses and IoU-based losses. Here, we recall them.

\subsection{Overview of $\ell_n$-norm losses and IoU-based losses}

The category of $\ell_n$-norm losses is mainly based on the $\ell_1$-norm and the $\ell_2$-norm which have some drawbacks. $\ell_1$ is less sensitive to outliers in the data, but is not differentiable at zero. Whereas $\ell_2$ is differentiable everywhere, but is highly sensitive to outliers. Therefore, Fast R-CNN \cite{Girshick_FastRCNN} and Faster R-CNN \cite{RenFasterRCNN} default use a \textit{Smooth} $\ell_1$ loss (originally defined as Huber loss \cite{Huber1964RobustEO}), which is differentiable everywhere and less sensitive to outliers than the $\ell_2$ loss used in the precedent object detection network, R-CNN \cite{GirshickRCNN}. A disadvantage of the \textit{Smooth} $\ell_1$ loss~\cite{Girshick_FastRCNN} is that it depends on a positive real parameter (controlling the transition from $\ell_1$ to $\ell_2$) that must be selected. Based on the \textit{Smooth} $\ell_1$ loss, there are other modifications, for example, \textit{Dynamic smooth} $\ell_1$ loss \cite{Zhang_DynamicRCNN} or \textit{Balanced} $\ell_1$ loss \cite{Pang_LibraRCNN_BalancedL1}. However, the main disadvantages of using the $\ell_n$-norm in general are ignoring the correlations between the four variables of the bounding boxes $(x; y;w; h)$, which is inconsistent with reality, and bias against large bounding boxes, which basically obtain large penalties in the calculation of the localization errors. 

The second category, IoU-based losses, jointly regresses all the bounding box variables as a whole unit; they are normalized and insensitive to the scales of the problem. The original IoU loss function \cite{IoU2016} was directly derived from the IoU metric. The main issue is that it does not respond to difference when the bounding boxes do not overlap; for such cases, the maximal loss value is produced. Meanwhile, a Generalized IoU (GIoU) loss \cite{GIoU2019} resolves the regression issue for the non-overlapping cases. However, both the IoU and the GIoU losses have a slow convergence. Distance IoU (DIoU) loss \cite{DIoU.CIoU2020} considers a normalized distance between the central points of the boxes. In addition, Complete IoU (CIoU) loss \cite{DIoU.CIoU2020} assumes three geometric components: the overlap area, the distance between the central points, and the aspect ratio. CIoU significantly improved the localization accuracy and convergence speed. However, the aspect ratio is not yet well defined. Based on CIoU, there are other modifications, for example, Improved CIoU (ICIoU) loss \cite{Wang_ImprovedCIoU2021} that utilizes the ratios of the corresponding widths and heights of the bounding boxes;
%a Scale-sensitive IoU (SIoU) loss \cite{Du_ScaleSensitiveIoU2021} adds a new geometric factor "area difference" between the boxes that affects the calculation of regression loss; 
Efficient IoU (EIoU) loss \cite{ZHANG_FocalandEfficientIoU2022} redefines the ratios of widths and heights between the boxes; A Focal EIoU loss \cite{ZHANG_FocalandEfficientIoU2022} was designed to improve the performance of the EIoU loss. Other IoU-based loss functions and improvements came, for example, with a SCYLLA IoU (SIoU) loss \cite{gevorgyan2022siou} where four cost functions are considered: the IoU cost, the angle cost, the distance cost and the shape cost; or a Balanced IoU (BIoU) loss \cite{Ravi_BIoU2022} where the parameterized distance between the centers and the minimum and maximum edges of the bounding boxes is addressed to solve the localization problem.

\subsection{Closer look on the selected IoU-based losses}
Intersection over Union (IoU) \cite{IoU2016} is a measure of comparison of the similarity between two arbitrary shapes $A,A' \subseteq \mathbb{S}\in \mathbb{R}^n$
\begin{equation}
     IoU = \frac{|A\cap A'|}{|A\cup A'|}.
\end{equation}
IoU as a similarity measure is independent of the space scale of $\mathbb{S}$ and can be transformed to the distance $(1-IoU)$ that satisfies all the standard metric properties. Therefore, it is popular for evaluating many 2D/3D computer vision tasks, mainly for image segmentation. However, IoU also has weaknesses, so it does not reflect different alignments of $A$ and $A'$ as long as their intersection is equal. Moreover, if there is no intersection between $A$ and $A'$, IoU is always zero and does not reflect any additional information, for example, the distance between $A$ and $A'$, their different sizes, areas, etc. 

In general, there is no simple and fast analytic solution to calculate the intersection between two arbitrary non-convex shapes. Fortunately, for the 2D object detection task, where the aim is to compare two axis-aligned bounding boxes, the solution is straightforward. Furthermore, IoU can be used directly as a loss function \cite{IoU2016}, i.e., $\mathcal{L}_{IOU} = 1 - IoU$ to optimize deep neural network-based object detectors. However, $\mathcal{L}_{IOU}$ still suffers from the weaknesses described above and its convergence speed is slow. Therefore, there are many modifications to the original IoU loss that aim to overcome its drawbacks and improve the accuracy of localization and convergence speed. Generally, IoU-based loss functions can be commonly defined as follows:
\begin{equation}\label{IoUbased}
     \mathcal{L}(B^g, B^p) = \mathcal{L}_{IOU}(B^g, B^p) + \mathcal{R}(B^g, B^p),
\end{equation}
where $\mathcal{L}_{IOU}(B^g, B^p) = 1 - IoU$ is the standard IoU loss between the ground truth $B^g$ and the predicted bounding box $B^p$. Then $\mathcal{R}(B^g, B^p)$ denotes the penalty term that specifies the particular modification. Our loss function proposed in Section~\ref{sec-smooth} also follows Formula (\ref{IoUbased}). In the following, we briefly describe the most commonly used IoU-based loss functions, which are further used for comparison.

The Generalized IoU (GIoU) loss \cite{GIoU2019} considers a minimum convex area $C$ that contains both boxes $B^g$ and $B^p$. The penalty term $\mathcal{R}(B^g, B^p)$ then provides a ratio of the area difference between the area $C$ and the union of the boxes. Therefore, the GIoU loss significantly enlarges the impact area by considering the non-overlapping cases of $B^g$ and $B^p$. Similarly to IoU, the GIou loss is invariant to the scale of the regression problem, and as a distance it is again a metric. But there are also drawbacks. In the cases at horizontal and vertical orientations it still carries large errors. The penalty term aims to minimize the area difference between $C$ and the union of $B^g$ and $B^p$, but this area is often small or zero (when two boxes have inclusion relationships), and then the GIoU loss almost degrades to the IoU loss. This yields a very slow convergence.

The Distance IoU (DIoU) loss \cite{DIoU.CIoU2020} minimizes a distance between $B^g$ and $B^p$. In particular, the penalty term $\mathcal{R}(B^g, B^p)$ provides a ratio of the Euclidean distance between the central points of the two boxes $B^g$ and $B^p$ and the diagonal length of a minimum convex area $C$ containing both boxes. It is again scale-invariant to the regression problem. In contrast to GIoU, for cases with inclusion of boxes $B^g$ and $B^p$, or in horizontal and vertical orientations, the DIoU loss can cause very fast convergence. However, in the case of inclusion and the central points of the boxes aligned with each other, the DIoU loss again degrades to the IoU loss. 

The Complete IoU (CIoU) loss \cite{DIoU.CIoU2020} considers three important geometric factors, i.e., the overlap area, the central point distance, and the aspect ratio. In this case, the penalty term $\mathcal{R}(B^g, B^p)$ is composed of the penalty term for DIoU (central point distance ratio) and the aspect ratio measured by a relationship of a width-to-height ratio of $B^g$ and a width-to-height ratio of $B^p$. The convergence speed and bounding box regression accuracy of the CIoU loss are significantly improved compared to the previous loss functions. However, the aspect ratio is not yet well defined. It just reflects the discrepancy of the width-to-height ratios between the two boxes, rather than the real relations between the corresponding widths and heights of the boxes. This can lead to cases where the width-to-height ratios of both boxes are the same, even when the predicted box is smaller or larger.

The SCYLLA IoU (SIoU) loss \cite{gevorgyan2022siou} considers four cost functions: IoU cost, angle cost, distance cost, and shape cost. In particular, the penalty term $\mathcal{R}(B^g, B^p)$ consists of a distance-based term and a shape-based term. The distance-based term works with the distance between the central points of the two boxes $B^g$, $B^p$ and the angles between the vector of central points and the axes $x$, $y$. The idea is first to bring the prediction $B^p$ to the closest axis $x$ or $y$ and then to continue the approach along the relevant axis. The shape-based term works with ratios of the corresponding widths and heights of the boxes $B^g$, $B^p$ and the degree of attention that should be paid to the shape cost. Since SIoU loss introduces the vector angle between the required regressions, it can accelerate the convergence speed and improve the accuracy of the regression. 

\section{Adding smoothing part to IoU loss}
\label{sec-smooth}

\begin{figure}[!h]
    \centering
    \subfigure[Ground truth bounding box.]{\includegraphics[width=29mm]{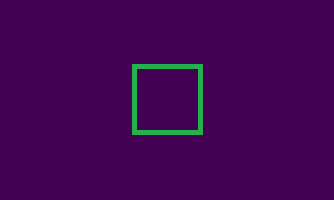}}
    \subfigure[Smoothing loss area constructed using the ground truth.]{\includegraphics[width=29mm]{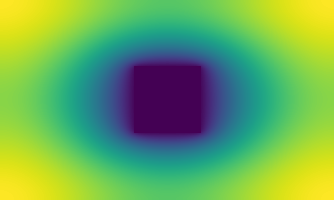}}
    \subfigure[Predicted boun\-ding box.]{\includegraphics[width=29mm]{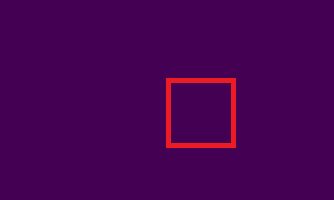}}
    \subfigure[Smoothing loss area corresponding to the predicted box.]{\includegraphics[width=29mm]{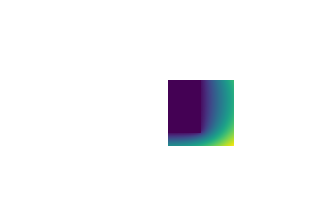}}
    \caption{2D visualization of how the smoothing part is constructed.
    The smoothing loss value is computed as the mean from the smoothing loss area (d). The formalism given in Section~\ref{sec-smooth} describes how such a process is carried out analytically. The collapsing of prediction into a small area inside the ground truth is prevented, primary by the fact that the loss inside the ground-truth part is zero, so the decreasing size of the predicted box does not decrease the loss, and secondary by adding the standard IoU part.} 
    \label{pic_smootnig_part}
\end{figure}

We assume $B=\{x_1, y_1, x_2, y_2\}$ to be a bounding box, where we use standard notation and consider $x_1, y_1$ to be the top-left point and $x_2, y_2$ the bottom-right point, together determining the rectangular area. In particular, we assume a couple of bounding boxes $B^g, B^p$ representing the ground truth bounding box and the predicted bounding box, respectively. The objective is to refine $B^p$ to match $B^g$ by minimizing the value of the corresponding loss function. We propose using a novel \textit{smoothing} version of the original IoU loss defined by the general formula (\ref{IoUbased}) and specified as follows:

\begin{equation}
     \mathcal{L_S}(B^g, B^p) = \mathcal{L}_{IOU}(B^g, B^p) + \mathcal{R_S}(B^g, B^p),
\end{equation}
where $\mathcal{L}_{IOU}(B^g, B^p)$ is the standard IoU loss \cite{IoU2016} as mentioned above and the penalty term $\mathcal{R_S}(B^g, B^p)$ denotes a \textit{smoothing} part of the loss that is the proposal of this study. The purpose of the smoothing part is to obtain resistance to noisy labels (similar to label smoothing~\cite{muller2019does}) and navigate the gradient to converge faster, so the motivation is similar to the other variants of losses based on IoU. 

The smoothing part $\mathcal{R_S}(B^g, B^p)$ is assumed to be a linear space with values that increase from the ground truth bounding box to the border of the input image, and thus cover the entire spatial search space; see \textbf{Fig. \ref{pic_smootnig_part}}. In particular, we consider the smoothing part to be specified by the offsets between the boxes $B^p$ and $B^g$. In addition, the offsets are loaded with weights designated by the distance between $B^p$ and $B^g$. The offsets and their weights are defined below.

Right offset:
\begin{equation}
    \underline{\mathbf{d}}^R = (\underline{x}^R, \underline{y}^R) = (\max (x^p_{1} - x^g_{2},0), \max (y^p_{2} - y^g_{1},0)),
\end{equation}

\begin{equation}
    \overline{\mathbf{d}}^R = (\overline{x}^R, \overline{y}^R) = (\max (x^p_{2} - x^g_{2},0), \max (y^p_{1} - y^g_{1},0)),
\end{equation}

\begin{equation}
    \mathbf{d}^R = (d^R_x, d^R_y) = \overline{\mathbf{d}}^R - \underline{\mathbf{d}}^R.
\end{equation}

Left offset:
\begin{equation}
    \underline{\mathbf{d}}^L = (\underline{x}^L, \underline{y}^L) = (\max (x^g_{1} - x^p_{1},0), \max (y^g_{2} - y^p_{2},0)),
\end{equation}

\begin{equation}
    \overline{\mathbf{d}}^L = (\overline{x}^L, \overline{y}^L) = (\max (x^g_{1} - x^p_{2},0), \max (y^g_{2} - y^p_{1},0)),
\end{equation}

\begin{equation}
    \mathbf{d}^L = (d^L_x, d^L_y) = \underline{\mathbf{d}}^L - \overline{\mathbf{d}}^L.
\end{equation}

Let $b^g_{c} = [x^g_c, y^g_c]$ be the central point of $B^g$ and
$$\mathbf{d}^c = (d^c_x, d^c_y) =(\max (x^g_c, 1 - x^g_c), \max (y^g_c, 1 - y^g_c)).$$
We define the weights of the right and left offsets,respectively, as follows:
\begin{equation}
    \mathbf{\omega}^R = (\omega^R_x, \omega^R_y) = \left(1 - \frac{\underline{x}^R + \overline{x}^R}{2d^c_x}, 1 - \frac{\underline{y}^R + \overline{y}^R}{2d^c_y}\right),
\end{equation}

\begin{equation}
    \mathbf{\omega}^L = (\omega^L_x, \omega^L_y) = \left(1 - \frac{\underline{x}^L + \overline{x}^L}{2d^c_x}, 1 - \frac{\underline{y}^L + \overline{y}^L}{2d^c_y}\right).
\end{equation}

Finally, $\mathcal{R_S}(B^g, B^p)$ is given by
\begin{equation}
    \mathcal{R_S}(B^g, B^p) = 1- \frac{\ell_{x} \ell_{y}}{4},
\end{equation}
where
\begin{equation}
    \ell_x = (1 - d^R_x)\omega^R_{x} + (1 - d^L_x)\omega^L_{x},
\end{equation}

\begin{equation}
    \ell_y = (1 - d^R_y)\omega^R_{y} + (1 - d^L_y)\omega^L_{y}.
\end{equation}

$\square$

\noindent The proposed loss function mimics the standard properties of a distance, namely:
\begin{enumerate}
    \item The proposed loss function $\mathcal{L_S}(B^g, B^p)$ is invariant to the scale of the regression problem.
    \item When the bounding boxes perfectly match, then $$\mathcal{L_S}(B^g, B^p) = \mathcal{L}_{IOU}(B^g, B^p) = 0.$$ When the boxes are far away, then $$\mathcal{L_S}(B^g, B^p) \to 2.$$
    \item The smoothing part $\mathcal{R_S}(B^g, B^p)$ is always non-negative and therefore, $$\mathcal{L_S}(B^g, B^p) \geq \mathcal{L}_{IOU}(B^g, B^p).$$
\end{enumerate}

\section{Experimental evaluation}
\label{sec-benchmark}
The current benchmarks are heavily dependent on the famous COCO dataset~\cite{lin2014microsoft}, which is huge and evaluates the entire object detector, including the classification and confidence parts. To omit the influence of these parts, we propose our own lightweight dataset consisting of 304 images, where each image includes exactly one object to be detected. The task is then solved by training a backbone that produces four values that are necessary to perform the regression.

The detailed setting of the experiment is as follows: 50\% fixed train/test split without data leak, fixed resolution of 512$\times$384px, backbone EfficientNetB2V2~\cite{tan2021efficientnetv2}, batch size 16, 6000 iterations, Adadelta optimizer~\cite{zeiler2012adadelta} with default LR (1.0) and the decrease to 0.4 and 0.1 after 3000$^{th}$ and 4500$^th$ iterations. Each of the tested losses was used in three separate runs, i.e., trained from scratch. 

\begin{figure}[!h]
    \centering
    \includegraphics[width=23.5mm]{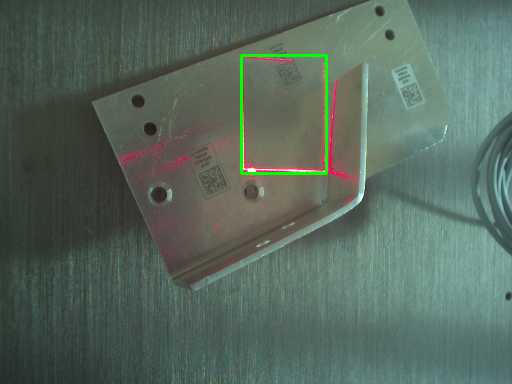}
    \includegraphics[width=23.5mm]{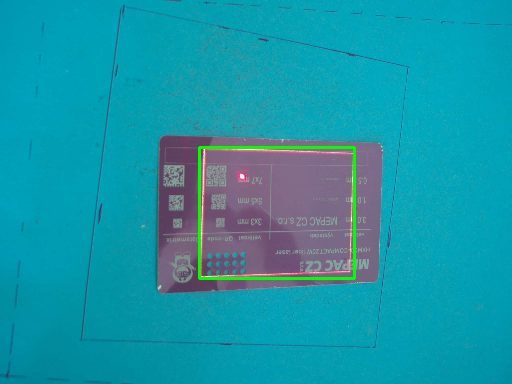}
    \includegraphics[width=23.5mm]{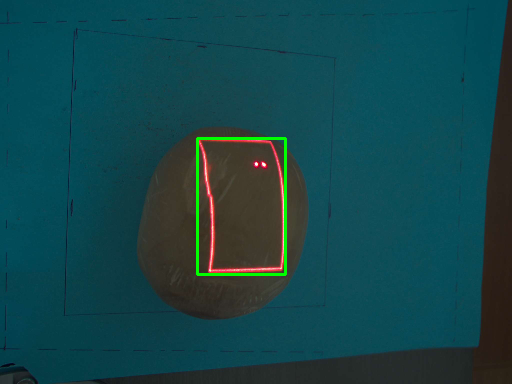}
    \includegraphics[width=23.5mm]{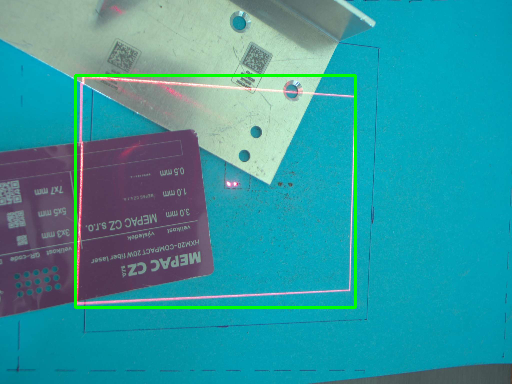}
    \includegraphics[width=23.5mm]{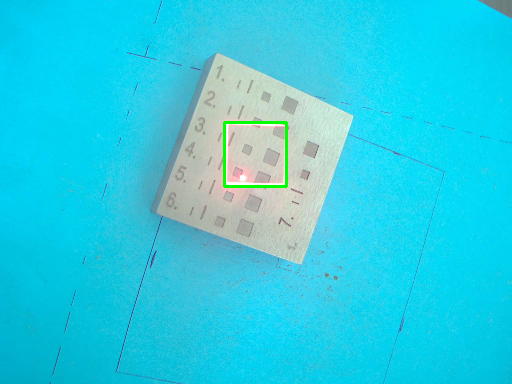}
    \caption{Images from the training dataset where the goal is to detect the red box. The ground truth label is visualized by the green color.} 
    \label{img-clean}
\end{figure}

\begin{figure}[!h]
    \centering
    \includegraphics[width=23.5mm]{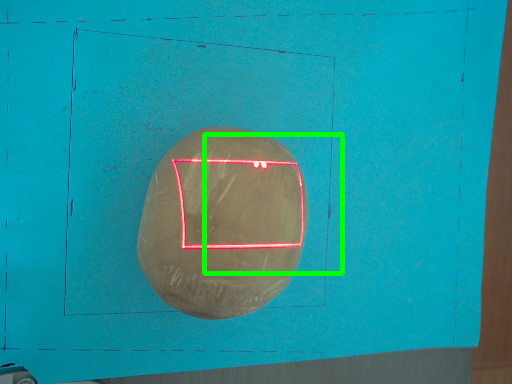}
    \includegraphics[width=23.5mm]{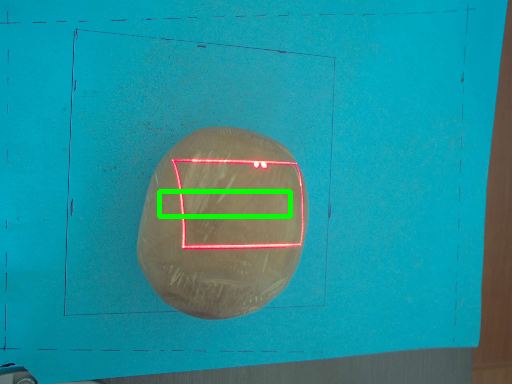}
    \includegraphics[width=23.5mm]{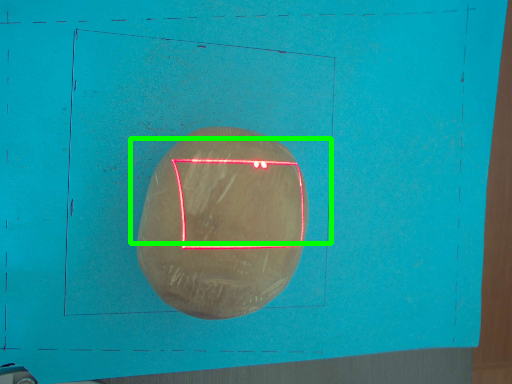}
    \includegraphics[width=23.5mm]{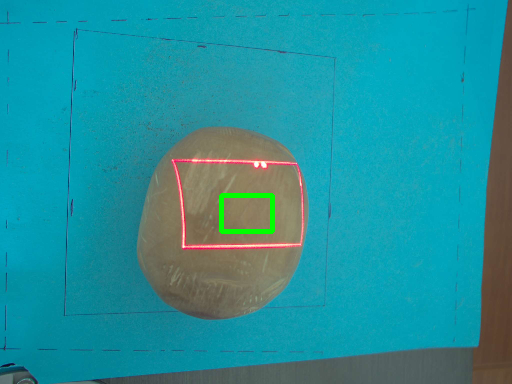}
    \includegraphics[width=23.5mm]{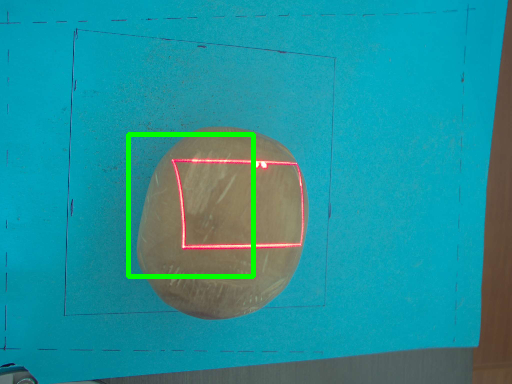}
    \caption{Images from the 40\% noisy training dataset where the goal is to detect the red box. The ground truth label is visualized by the green color.} 
    \label{img-noisy}
\end{figure}

The detailed results are shown in Table~\ref{results-clean} for the original dataset and in Tables~\ref{results-noise20}-\ref{results-noise60} for synthetically added noise of 20-60\% of the side size of the bounding box. For an illustration of the images, see Figures~\ref{img-clean} and \ref{img-noisy} for the clean and noisy version.  The notation$\pm$ used in the tables expresses the difference between the best and the worst run. Noise means that each coordinate is added to $n \sim \mathbb{U}(-\mu s, \mu s)$ where $\mu \in [0,1]$ is the noise level and $s$ corresponding side size; width for the $x$ coordinates and height for the $y$ coordinates.

\noindent The interpretation of the results and highlights is as follows:
\vspace{-2mm}
\begin{itemize}
    \item Smoothing IoU yields the best results on both clean and noisy dataset.
    \item Smoothing IoU achieves the lowest overfit on clean dataset.
    \item Smoothing IoU achieves the largest underfit on noisy dataset, i.e., it is least sensitive to noisy labels among the benchmarked IoU variants.
    \item With an increase in the noise level, the regression accuracy on testset remains stable for smoothed IoU while strongly decreasing for other IoU variants.
    \item The proposed loss function is so superior that, when trained on 40\% noisy dataset, it still obtains a better test accuracy than the other losses trained on a clean dataset.
\end{itemize}

\begin{table}[!ht]
\centering
\caption{Clean dataset}\label{results-clean}
\begin{tabular}{p{25mm}| c c c| c c }
\hline
&\multicolumn{3}{c}{Measured IOU similarity}\\
Loss type&~~~Train avg~~~&~~~Test avg~~~&~~~Test best~~~&~~~Test $\pm$~~~&~~~Overfit~~~\\
\hline
IOU &\cellcolor{c_red}0.635 &0.536 &0.559 &\cellcolor{c_green}0.051&0.099\\
SIOU~\cite{gevorgyan2022siou} &0.668 &\cellcolor{c_red}0.378 &\cellcolor{c_red}0.495&\cellcolor{c_red}0.270&\cellcolor{c_red}0.290 \\
DIOU~\cite{DIoU.CIoU2020} &0.668 &0.504&0.615&  0.228& 0.164 \\
CIOU~\cite{DIoU.CIoU2020}& 0.656&0.526 &0.579 &0.082&0.130 \\
IOU + smooth &\cellcolor{c_green}0.713 &\cellcolor{c_green}0.684 &\cellcolor{c_green}0.729&0.073&\cellcolor{c_green}0.029 \\
\hline
\end{tabular}
\end{table}

\begin{table}[!ht]
\centering
\caption{Noisy dataset, 20\%. Because the dataset is noisy, train avg and overfit are informative only.}\label{results-noise20}
\begin{tabular}{p{25mm}| c c c| c c }
\hline
&\multicolumn{3}{c}{Measured IOU similarity}&&\\
Loss type&~~~Train avg~~~&~~~Test avg~~~&~~~Test best~~~&~~~Test $\pm$~~~&~~~Overfit~~~\\
\hline
IOU &0.509& 0.471& 0.555 &\cellcolor{c_red}0.138& ~0.038\\
SIOU~\cite{gevorgyan2022siou} &0.551&0.550&0.591&\cellcolor{c_green}0.064& ~0.001\\
DIOU~\cite{DIoU.CIoU2020} &0.521 &\cellcolor{c_red}0.442 &\cellcolor{c_red}0.482& 0.079 & ~0.113\\
CIOU~\cite{DIoU.CIoU2020} &0.534 & 0.494& 0.532&0.087&~0.040 \\
IOU + smooth &0.535&\cellcolor{c_green}0.588 &\cellcolor{c_green}0.628& 0.090& -0.053\\
\hline
\end{tabular}
\end{table}

\begin{table}[!ht]
\centering
\caption{Noisy dataset, 40\%. Because the dataset is noisy, train avg and overfit are informative only.}\label{results-noise40}
\begin{tabular}{p{25mm}| c c c| c c }
\hline
&\multicolumn{3}{c}{Measured IOU similarity}&&\\
Loss type&~~~Train avg~~~&~~~Test avg~~~&~~~Test best~~~&~~~Test $\pm$~~~&~~~Overfit~~~\\
\hline
IOU &0.340 &\cellcolor{c_red}0.386 &\cellcolor{c_red}0.453 & 0.112 & -0.046\\
SIOU~\cite{gevorgyan2022siou} &0.399& 0.444& 0.562&\cellcolor{c_red}0.232& -0.045\\
DIOU~\cite{DIoU.CIoU2020} &0.363& 0.442& 0.530& 0.163& -0.079\\
CIOU~\cite{DIoU.CIoU2020} &0.383& 0.423& 0.507& 0.166& -0.040\\
IOU + smooth & 0.387& \cellcolor{c_green}0.546&\cellcolor{c_green}0.574&\cellcolor{c_green}0.054& -0.159\\
\hline
\end{tabular}
\end{table}

\begin{table}[!ht]
\centering
\caption{Noisy dataset, 60\%. Because the dataset is noisy, train avg and overfit are informative only.}\label{results-noise60}
\begin{tabular}{p{25mm}| c c c| c c }
\hline
&\multicolumn{3}{c}{Measured IOU similarity}&&\\
Loss type&~~~Train avg~~~&~~~Test avg~~~&~~~Test best~~~&~~~Test $\pm$~~~&~~~Overfit~~~\\
\hline
IOU &0.202& \cellcolor{c_red}0.294& 0.362& \cellcolor{c_red}0.156& -0.092\\
SIOU~\cite{gevorgyan2022siou} &0.271& 0.345& 0.378& 0.086& -0.074\\
DIOU~\cite{DIoU.CIoU2020} &0.250& 0.328& \cellcolor{c_red}0.347& \cellcolor{c_green}0.040& -0.078\\
CIOU~\cite{DIoU.CIoU2020} &0.258& 0.349& 0.417& 0.149& -0.091\\
IOU + smooth &0.241& \cellcolor{c_green}0.509& \cellcolor{c_green}0.581& 0.107& -0.268\\
\hline
\end{tabular}
\end{table}

\section{Summary}
In this contribution, we have designed the smoothing modification of the standard IoU loss function for the bounding box regression. The proposed smoothing part navigates the gradient through the whole image to converge faster. The exact analytical formalism of the smoothing part is also described in order to be simply integrated to the standard architectures. Based on experimental evolution, we show that the smoothing version of IoU outperforms the benchmarks IoU, SIoU, DIoU, and CIoU. Moreover, the proposed smoothing IoU loss function is resistant to noisy labels. In comparison with the mentioned benchmarks where the regression accuracy strongly decreases for increasing noisy labels, the smoothing IoU remains stable. In industrial applications where noisy labels often appear and therefore robustness is required, the simple implementation of the smoothing IoU loss is the great advantage. The reference implementation is available at gitlab.com/irafm-ai/smoothing-iou.

\bibliographystyle{unsrt}  
\bibliography{object_detection}  %%% Remove comment to use the external .bib file (using bibtex).


\end{document}